\documentclass[lettersize,journal]{IEEEtran}

\usepackage{amsmath,amssymb,amsfonts}
\usepackage{algorithmic}
\usepackage{algorithm}
\usepackage{array}
\usepackage{booktabs}
\usepackage{multirow}
\usepackage{enumitem}
\usepackage{textcomp}
\usepackage{stfloats}
\usepackage{url}
\usepackage{graphicx}
\usepackage{cite}
\usepackage[hidelinks]{hyperref}

\newtheorem{definition}{Definition}

\newcommand{\codeurl}{https://github.com/shabanpour/latent-channel-completion}

\begin{document}

\title{Latent Channel Completion for Forecasting in\\
Sensor Networks with Failed Detectors}

\author{Mehran~Shabanpour$^{*,\dag}$, Sadaf~Khademi$^{*,\dag}$, Maryam Hosseini$^{\dag}$, Haiyan Xu$^{\ddag}$, \\\vspace{.04in} Konstantinos N. Plataniotis$^{\ddag}$, Arash Mohammadi$^{\dag}$
\thanks{ ${^\dag}$  M. Shabanpour, S. Khademi, M. Hosseini, and A. Mohammadi are with the I-SIP Lab at Cybersecurity \& Intelligent Systems Engineering (CISE), Concordia University, Montreal, QC,
Canada. ${^*}$ Authors contributed equally to this work.\\
${^\ddag}$  H. Xu and K.N. Plataniotis are with the Multimedia Lab at Electrical \& Computer Engineering (ECE) Department, University of Toronto, Toronto, ON, Canada.}
\thanks{This work was partially supported by Natural Sciences \& Eng. Research Council (NSERC) of Canada; NSERC Discovery Grant RGPIN-2023-05654.}
\thanks{An implementation of LCC, together with the failure protocol and the scripts that reproduce every table and figure in this paper, is available at \url{\codeurl}.}}
\markboth{IEEE Transactions on Reliability, Special Section on Trustworthy and Reliable AI}%
{Shabanpour \MakeLowercase{\textit{et al.}}: Latent Channel Completion for Forecasting in Sensor Networks with Failed Detectors}

\maketitle

\begin{abstract}
Monitoring networks for traffic and utility infrastructure lose sensors
routinely, and forecasting models that couple sensors together to gain accuracy
also inherit an exposure to those losses. 
When a detector stops reporting, the
substituted value is embedded as though it were a reading and propagates through
the shared representation that couples the network, degrading forecasts for
sensors that are still healthy. Because these models are selected on benchmarks
where every sensor reports, the exposure is absent from the numbers used to
choose them.
We introduce Latent Channel Completion, a channel-axis
joint-embedding predictive objective that reconstructs missing sensors' latent
representations rather than their values, since a coupling operator consumes
embeddings and never sees raw readings. Channels are masked during training and
their representations predicted from surviving channels against stop-gradient
targets; at inference the operator fills in whichever sensors the monitoring
system reports as dark, using parameters shared with the forecaster. On three
highway detector networks, after half of each network goes dark, the accuracy
advantage of coupling changes by $-1.0$, $-0.5$ and $+8.8$ percentage points,
against losses of $10.2$ to $50.3$ points for the unmodified forecaster, which
on one network falls behind an uncoupled reference entirely. Against
cross-channel ridge imputation drawing on the same surviving sensors, completing
the latent lowers error by a further $13.9\%$ and $28.6\%$, and on complete
inputs, the method matches its backbone. The benefit tracks how much a network
gains from coupling when intact, giving operators a criterion evaluable before
any outage occurs.
\end{abstract}

\begin{IEEEkeywords}
Deep learning, fault tolerance, missing data, network reliability, self-supervised
learning, sensor networks, time series forecasting.
\end{IEEEkeywords}

\section{Introduction}
\label{sec:intro}

\IEEEPARstart{T}{raffic} management centers use networks of roadway
detectors to generate short-horizon forecasts that support ramp
metering, incident response, and traveler information. Meanwhile, operators' role is shifting toward supervising automated
decision-making pipelines rather than making each decision
individually~\cite{wang2021perspectives}.
However, detector networks are vulnerable to disruptions when loop sensors degrade, communication links fail, or power outages interrupt measurements across entire network segments. Consequently, a monitoring system covering hundreds of locations may operate with incomplete observations. 
In this setting, the operational challenge extends beyond forecasting
traffic at locations with missing measurements; it also requires
maintaining accurate forecasts at locations where sensors remain
operational.

Treating each series independently discards the spatial structure that helps make a detector
network predictable. 
In contrast, models that exploit cross-sensor correlation through
graph convolution over road adjacency~\cite{yu2018stgcn},
or by pooling all sensors into a shared representation, recover
accuracy that independent modeling leaves behind. Forecasting models for these networks increasingly rely on
joint representations of sensor measurements. 
On the detector networks studied here, joint modeling reduces mean
squared error by $39\%$--$63\%$ relative to forecasting each sensor's measurements independently~\cite{han2024softs}.

However, this coupling introduces a risk that evaluation of a complete-network does not reveal. When a sensor stops reporting, the monitoring pipeline substitutes a default value, the model embeds that substitute as a valid observation, and the shared representation computed across sensors can be distorted. This corrupted representation can propagate to predictions at other locations, including those whose sensor measurements remain unchanged. Thus, the same cross-sensor dependencies that improve forecasting accuracy can also expose operational sensors to errors induced by missing inputs elsewhere in the network. This effect is not captured by standard benchmark analyses that evaluate models with all sensors available.

We address this problem with Latent Channel Completion (LCC), a channel-axis joint-embedding predictive objective that reconstructs missing sensors' latent representations rather than their raw measurements. This design targets the point at which
the corruption enters. A coupling operator never sees a sensor's raw window; it consumes the sensor's embedding. Supplying that embedding directly is a lower-dimensional problem than reconstructing the readings that would have produced it, and it is the problem the architecture actually poses. LCC learns to predict a channel's embedding from the surviving channels by masking channels during training and predicting their representations against stop-gradient targets, then invokes that operator at inference for whichever sensors the monitoring system reports as dark. Fig.~\ref{fig:overview} illustrates the proposed framework.

Across three detector networks (PEMS03, PEMS04, and PEMS08), the accuracy advantage of coupling survives the loss of half the network. On PEMS03, the advantage over independent forecasting falls from $42.9\%$ to $-7.4\%$ for the unmodified coupled model as sensors fail, while LCC holds $42.0\%$; the pattern repeats on PEMS04 and PEMS08 and across forecast horizons from $12$ to $48$ steps. 
Compared with cross-channel ridge imputation, a repair that reconstructs the missing readings from the surviving ones and uses the same information as LCC, completing the latent representation lowers error by a further $13.9\%$ on PEMS03 and $28.6\%$ on PEMS04. On complete inputs, LCC matches its backbone's forecasting
accuracy, providing robustness to sensor failure without an
observed accuracy penalty during normal operation.

The main contributions of this work are summarized as follows:

\begin{itemize}
\item An analysis of how sensor failures propagate through shared
representations and degrade forecasts at operational sensors,
motivating the completion of missing embeddings before
cross-sensor mixing.

\item Latent Channel Completion, a channel-axis predictive objective with a
dual-path training scheme that keeps complete-input forecasting in distribution
while training the completion route through shared parameters. To the best of our
knowledge, latent prediction has not previously been applied across the channel
axis as a mechanism for handling sensor loss at inference.

\item Evidence that the coupling advantage becomes invariant to sensor failure.
After half the sensors go dark, LCC's advantage changes by $-1.0$, $-0.5$ and
$+8.8$ percentage points on the three detector networks, against losses of
$10.2$ to $50.3$ points for the unmodified model. Moreover, an ablation using fixed trained weights attributes
the effect to inference-time completion rather than to masked
training alone.

\item A deployment criterion for when the method applies. The advantage LCC
retains tracks the advantage a network offers when intact, measured across five
networks spanning $0.7\%$ to $62.8\%$. Thus, an operator can determine whether
completion is worth deploying by comparing coupled and independent forecasters
once on complete data.
\end{itemize}

The remainder of this paper is organized as follows:
Section~\ref{sec:related} reviews cross-channel forecasting,
robustness evaluation, missing-data recovery, and latent predictive
objectives. Section~\ref{sec:problem} formulates the forecasting
problem under declared sensor failure.
Section~\ref{sec:method} presents LCC,
Section~\ref{sec:setup} describes the experimental setup, and
Section~\ref{sec:results} reports the results.
Section~\ref{sec:discussion} discusses deployment implications,
threats to validity, and limitations.
Section~\ref{sec:conclusion} concludes the paper.

\section{Related Work}
\label{sec:related}

\subsection{Cross-Channel Modeling in Multivariate Forecasting}
\label{subsec:rw_channel}

Multivariate forecasters can be categorized by whether a channel's prediction depends on
the other channels. Channel-independent models forecast each series from its own
history and proved unexpectedly strong on standard
benchmarks~\cite{nie2023patchtst}, while channel-dependent models~\cite{moemba2025, shabanpour2026decode} recover the
cross-series structure that independence discards. 
In spatially organized sensor networks, that structure is substantial. 
Graph convolutional forecasters exploit road-network adjacency~\cite{yu2018stgcn}, while recent work extends the idea to irregular detector layouts~\cite{duan2026patchstg} and large road networks through learned clustering~\cite{ali2026emagn}. Alternatively, aggregation
architectures avoid an explicit graph by pooling all channels into a shared
representation that is dispatched back to each series~\cite{han2024softs},
which is the mechanism this paper builds on.

The trade-off between the two families is itself an active question. Several
recent methods route dependencies selectively rather than committing to either
extreme, using spectral decomposition of a channel-patch
graph~\cite{qiu2026routing}, patch-specific spatial-temporal
filtration~\cite{hu2025timefilter}, or permutation-invariant relational
interaction that mitigates the overfitting channel-dependent models exhibit on
weakly coupled data~\cite{shen2026cpiri}. These methods weigh cross-channel
information against its statistical cost on clean data. The cost this paper
measures is different in kind, as dependence on other sensors is also an exposure
to their failure, and that exposure is invisible to any evaluation conducted with
a complete network.

\subsection{Sensor-Fault Robustness Evaluation}
\label{subsec:rw_robustness}

Forecasting models are ranked almost entirely by nominal accuracy, and a body of
recent work argues that this ranking does not survive deployment conditions.
TS-Fault evaluates forecasters against structural faults rather than a single
error number~\cite{tsfault2026}, and a parallel benchmark studies cyber-physical
forecasting under noisy, biased, missing, and misaligned sensor
streams~\cite{cpsfault2026}. Analyses of spatio-temporal benchmarks reach a
compatible conclusion about how narrowly these models are
assessed~\cite{audit2026}. Outside forecasting, the same question has been posed
for multi-modal perception, where fusion mechanisms that outperform
single-modality baselines prove vulnerable once a sensor
fails~\cite{bev2026}.
This literature establishes that the problem exists and quantifies its extent. It
stops there by design; a benchmark measures fragility rather than removing it,
and prior studies propose no repair mechanism. The present work takes the
measurement as given and examines what architectural change restores the accuracy lost under sensor failure.

\subsection{Missing-Data Recovery for Monitoring Networks}
\label{subsec:rw_imputation}

Reconstructing absent sensor readings is a well-established field with strong deep
baselines. Bidirectional recurrent imputation~\cite{cao2018brits} and
self-attention imputation~\cite{du2023saits} recover missing values from both
temporal continuity and cross-variable correlation, and diffusion models have
since been applied to the same task~\cite{fadti2025, wang2024imputationsurvey}. Domain-specific variants address the
failure patterns that real networks exhibit, including communication-aware
recovery for distribution grids~\cite{grid2026} and blackouts that are missing
not-at-random because the outage depends on unobserved
conditions~\cite{blackout2026}. A separate line avoids reconstruction entirely,
either by modeling observational existence alongside
values~\cite{existence2026} or by handling irregular sampling without an
imputation stage~\cite{imputationfree2025}. Training-time corruption is also
effective, as exogenous dropout is a simple and strong baseline for covariate
corruption at inference~\cite{exodropout2026}.
These methods reconstruct the signal. A channel-dependent forecaster, however,
consumes a channel's embedding rather than its readings; reconstructing the
signal therefore solves a harder estimation problem than the architecture requires. Our
comparison in Section~\ref{subsec:imputation} holds the available information
fixed and varies only where the repair is applied, which isolates that
distinction.

\subsection{Latent Predictive Self-Supervision}
\label{subsec:rw_jepa}

Joint-embedding predictive architectures (JEPA) learn by predicting the representation
of masked content rather than reconstructing the content
itself~\cite{assran2023ijepa,bardes2024vjepa}, an objective that avoids spending
capacity on unpredictable detail. These ideas have also been rapidly adapted to time-series settings.
Latent prediction has been used for representation learning without augmentation
or masking~\cite{cfjepa2026,lenepa2026}, for anomaly prediction where the target
is a precursor state~\cite{scjepa2026}, for horizon-conditioned event
prediction under label scarcity~\cite{hepa2026}, and for physics-informed latent
world models of coupled variables~\cite{physjepa2026}. Domain-specific
formulations extend the idea to physiological recordings~\cite{eegjepa2026}, and
recent analyses examine the role of the encoder and the variational structure of
the objective~\cite{siamjepa2026,varjepa2026}.

These formulations predict along the temporal axis, masking spans of time and
predicting their representations from surrounding context; the objective serves
pretraining, and evaluation assumes complete inputs. This paper keeps the
predictive objective and changes the axis it operates on. Masking is applied
across channels rather than across time, and the completion operator is invoked
at inference rather than discarded after pretraining, which turns a
representation-learning objective into a mechanism for tolerating sensor loss.

\section{Problem Formulation}
\label{sec:problem}

\subsection{Monitoring Model and Notation}
\label{subsec:notation}

A monitoring network consists of $C$ sensors indexed by
$\mathcal{C}=\{1,\dots,C\}$, each reporting a scalar reading at a fixed
sampling interval. Over a window of $L$ consecutive steps the network produces
$\mathbf{X}\in\mathbb{R}^{L\times C}$, whose $c^{\text{th}}$ column
$\mathbf{x}_c\in\mathbb{R}^{L}$ is the history of sensor $c$. The forecasting
task is to predict the next $H$ steps for every sensor,
$\mathbf{Y}\in\mathbb{R}^{H\times C}$.

A channel-dependent forecaster couples the sensors. It first embeds each
channel independently through a shared map
$\phi:\mathbb{R}^{L}\!\rightarrow\!\mathbb{R}^{D}$, giving
$\mathbf{h}_c=\phi(\mathbf{x}_c)$ and the embedding matrix
$\mathbf{H}=[\mathbf{h}_1,\dots,\mathbf{h}_C]^{\!\top}\in\mathbb{R}^{C\times D}$.
A mixing operator $\pi$ then summarizes the network state into a shared
representation that is redistributed to every channel. In the aggregate-and-dispatch
family used throughout this work~\cite{han2024softs}, $\pi$ pools $\mathbf{H}$
into a single core $\mathbf{s}\in\mathbb{R}^{D}$ and concatenates it back to each
channel embedding, making the forecast for sensor $c$
dependent on the readings of all other sensors. A channel-independent forecaster omits $\pi$ entirely and
predicts each column of $\mathbf{Y}$ from the corresponding column of
$\mathbf{X}$ alone~\cite{nie2023patchtst}. Table~\ref{tab:notation} collects the
symbols used throughout.

\subsection{Sensor Failure at Inference}
\label{subsec:failure}

Deployed monitoring networks lose sensors. Detectors fail, communication links
drop, and power interruptions remove contiguous blocks of a network at
once~\cite{blackout2026,grid2026}. We model a failure episode
by a set $\mathcal{F}\subset\mathcal{C}$ of sensors that report no data over the
prediction window, with the surviving sensors
$\mathcal{A}=\mathcal{C}\setminus\mathcal{F}$. The sensor-failure rate
$p=|\mathcal{F}|/|\mathcal{C}\setminus\mathcal{E}|$ measures the extent of an
episode, where $\mathcal{E}$ is defined below.

Operationally, a failed sensor is not simply absent. The monitoring pipeline must
supply some value for it, and the standard fallback substitutes a statistic
computed on training data, typically the channel mean. Writing
$\mathrm{sub}(\cdot)$ for this substitution, the forecaster receives
$\tilde{\mathbf{X}}$ with
$\tilde{\mathbf{x}}_c=\mathrm{sub}(\mathbf{x}_c)$ for $c\in\mathcal{F}$ and
$\tilde{\mathbf{x}}_c=\mathbf{x}_c$ otherwise.

Two properties of this setting shape the evaluation. First, a forecast issued for
a sensor that is dark cannot be validated against ground truth in the field, and
an operator cannot act on it. Second, the quantity that matters to the operator is
what happens to the sensors that are still reporting. We therefore designate a
fixed subset $\mathcal{E}\subset\mathcal{C}$ of evaluation channels that
never fail, draw $\mathcal{F}$ only from $\mathcal{C}\setminus\mathcal{E}$, and
measure error on $\mathcal{E}$ alone. Because $\mathcal{E}$ is fixed across all
failure rates and all methods, errors are directly comparable along $p$. Therefore, any
change reflects the model's response to losing inputs, not a change in which
sensors are being scored.

\begin{definition}[Forecasting under declared sensor failure]
\label{def:task}
Given a window $\mathbf{X}\in\mathbb{R}^{L\times C}$, a failure set
$\mathcal{F}\subset\mathcal{C}\setminus\mathcal{E}$ known at inference time, and
the substituted input $\tilde{\mathbf{X}}$, learn
$f_\theta:\mathbb{R}^{L\times C}\times 2^{\mathcal{C}}\!\rightarrow\!\mathbb{R}^{H\times C}$
minimizing the error on the evaluation channels,
\begin{equation}
\label{eq:objective}
\mathbb{E}\!\left[\,\big\|\, f_\theta(\tilde{\mathbf{X}},\mathcal{F})_{:,\mathcal{E}}
- \mathbf{Y}_{:,\mathcal{E}} \,\big\|_2^2\,\right],
\end{equation}
for failure sets ranging from $\mathcal{F}=\varnothing$ to
$\mathcal{F}=\mathcal{C}\setminus\mathcal{E}$.
\end{definition}

\subsection{Assumptions}
\label{subsec:assumptions}

\begin{enumerate}[label=\textbf{A\arabic*.},leftmargin=2.6em]
\item \textbf{Failures are declared.} The identity of the failed set
$\mathcal{F}$ is available at inference. Monitoring systems track sensor health
independently of the forecaster, and a dark sensor is detectable from the absence
of transmission rather than inferred from its values.
\item \textbf{Failures persist across the window.} A sensor in $\mathcal{F}$
reports nothing for the whole of the $L$ steps. Section~\ref{sec:results}
relaxes this to partial outages in which a sensor fails part-way through the
window and leaves usable history behind it.
\item \textbf{The surviving network remains informative.} At least one sensor
outside $\mathcal{E}$ continues to report whenever $p<1$. This property holds by
construction of the failure model and is stated explicitly to ensure that the completion operator
of Section~\ref{sec:method} always has a non-empty context.
\end{enumerate}

Assumption~\textbf{A1} is what separates this setting from blind imputation. A forecaster
that knows which inputs are untrustworthy can treat them differently from
the ones it trusts, and the method developed next uses exactly that information.

\begin{table}[!t]
\centering
\caption{Summary of notation, grouped by role, with the equation in which each
quantity is defined or first used.}
\label{tab:notation}
\setlength{\tabcolsep}{4pt}
\renewcommand{\arraystretch}{1.15}
\begin{tabular}{@{}llc@{}}
\toprule
Symbol & Meaning & Defined in \\
\midrule
\multicolumn{3}{@{}l}{\emph{Network and data}}\\
$\mathcal{C}$, $C$      & sensor index set and its size            & --- \\
$L$, $H$, $D$           & lookback, horizon, embedding width       & --- \\
$\mathbf{X}$            & input window, $\mathbb{R}^{L\times C}$   & --- \\
$\mathbf{x}_c$          & history of sensor $c$                    & --- \\
$\mathbf{Y}$            & forecast target, $\mathbb{R}^{H\times C}$ & --- \\
\addlinespace[2pt]
\multicolumn{3}{@{}l}{\emph{Forecaster}}\\
$\phi$                  & per-channel embedding map                & --- \\
$\mathbf{h}_c$          & embedding of sensor $c$, $\phi(\mathbf{x}_c)$ & --- \\
$\mathbf{H}$            & stacked embeddings, $\mathbb{R}^{C\times D}$  & --- \\
$\pi$, $\mathbf{s}$     & coupling operator and shared representation & \eqref{eq:corrupt} \\
$\psi$                  & coupling stack and forecasting head      & \eqref{eq:loss} \\
$f_\theta$              & forecaster under declared failure        & \eqref{eq:objective} \\
\addlinespace[2pt]
\multicolumn{3}{@{}l}{\emph{Failure model}}\\
$\mathcal{F}$, $\mathcal{A}$ & failed and surviving sensors        & --- \\
$\mathcal{E}$           & evaluation sensors, which never fail     & \eqref{eq:objective} \\
$p$                     & sensor-failure rate over the pool        & --- \\
$\mathrm{sub}(\cdot)$   & substitution applied to a dark sensor    & --- \\
$\tilde{\mathbf{X}}$    & window after substitution                & \eqref{eq:objective} \\
\addlinespace[2pt]
\multicolumn{3}{@{}l}{\emph{Completion}}\\
$\mathbf{m}$, $\mathcal{M}$ & unavailability indicator and its support & \eqref{eq:mask} \\
$\mathbf{q}$            & learned mask vector                      & \eqref{eq:mask} \\
$\mathbf{H}^{\mathbf{m}}$ & masked embeddings                      & \eqref{eq:mask} \\
$g$, $\hat{\mathbf{H}}$ & completion operator and its predictions  & \eqref{eq:complete} \\
$\mathbf{H}^{\star}$    & embeddings after completion              & \eqref{eq:complete} \\
$\mathrm{sg}[\cdot]$    & stop-gradient                            & \eqref{eq:aux} \\
$r$, $\lambda$          & masking probability, auxiliary weight    & \eqref{eq:loss} \\
\addlinespace[2pt]
\multicolumn{3}{@{}l}{\emph{Evaluation}}\\
$\ell$                  & squared error over the horizon           & \eqref{eq:loss} \\
$\mathrm{MSE}$          & error on the evaluation sensors          & \eqref{eq:mse} \\
$A$                     & mixing advantage                         & \eqref{eq:advantage} \\
\bottomrule
\end{tabular}
\end{table}

\begin{figure*}[!t]
\centering
\includegraphics[width=\textwidth]{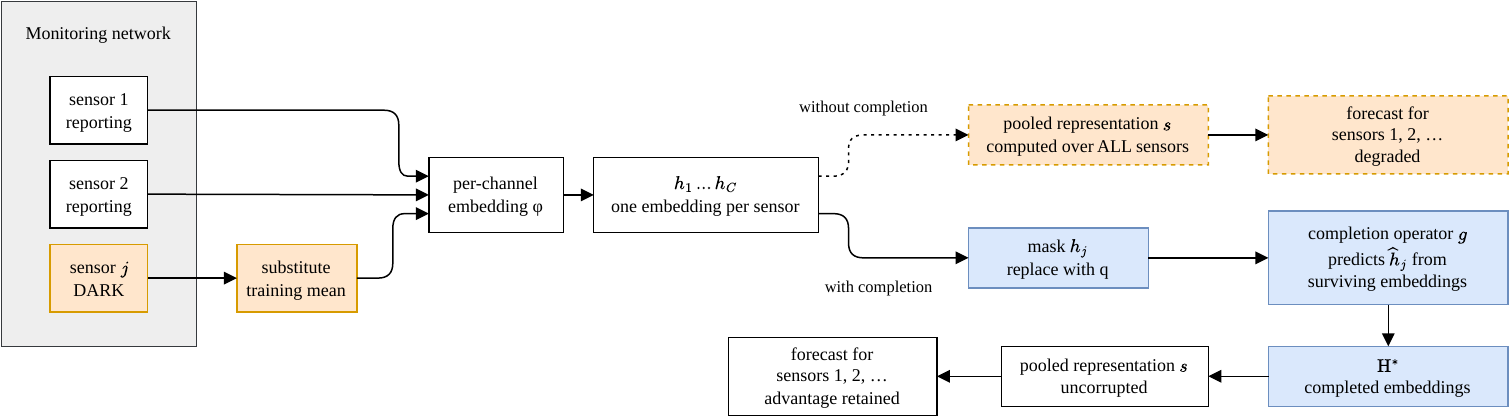}
\caption{Sensor failure in a coupled forecaster, and where completion
intercepts it. A dark sensor is replaced by a training statistic, embedded as
though it were a reading, and absorbed into the pooled representation that every
channel receives, degrading forecasts for sensors whose own data is intact
(dashed route). Latent Channel Completion predicts the missing embedding from
the surviving ones before pooling (solid route); the pooling operator and
forecasting head are shared between the two.}
\label{fig:overview}
\end{figure*}

\begin{figure*}[!t]
\centering
\includegraphics[width=\textwidth]{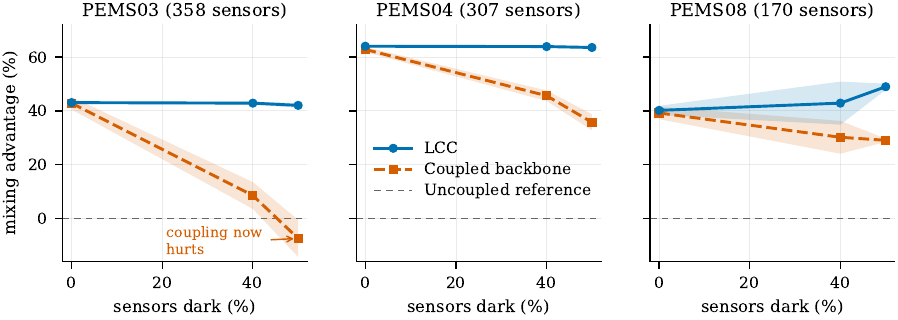}
\caption{Accuracy advantage of channel coupling as detectors go dark, measured
against a channel-independent reference (dashed line at zero). Bands span the
two seeds. The coupled backbone loses its advantage as sensors fail and on
PEMS03 falls below the uncoupled reference, while the same backbone with latent
completion holds its advantage across the range.}
\label{fig:degradation}
\end{figure*}

\begin{figure}[!t]
\centering
\includegraphics[width=\columnwidth]{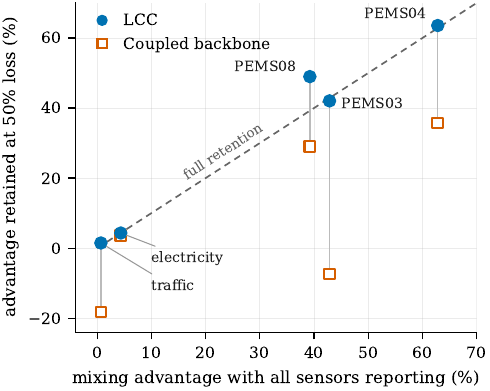}
\caption{Advantage retained after half the network goes dark, against the
advantage the network offers with every sensor reporting. Points on the diagonal
retain their full advantage. Vertical segments join the two systems on each
network; the five networks span coupling advantages from $0.7\%$ to $62.8\%$.}
\label{fig:tracking}
\end{figure}

\begin{figure}[!t]
\centering
\includegraphics[width=\columnwidth]{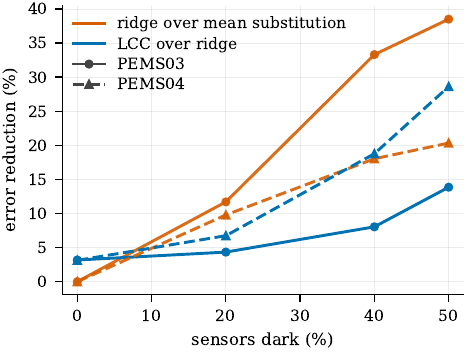}
\caption{Error reduction from repairing declared failures, relative to the
substitution applied downstream of each repair. Reconstructing readings with a
cross-channel ridge improves substantially on mean substitution, and completing
the embedding improves further on the ridge; both margins grow as more detectors
fail.}
\label{fig:repair}
\end{figure}

\section{Latent Channel Completion}
\label{sec:method}

Latent Channel Completion (LCC) is a channel-axis joint-embedding predictive
objective that reconstructs missing sensors' latent representations rather than
their values, addressing Definition~\ref{def:task}. This section first identifies where a channel-mixed forecaster
loses accuracy when sensors fail (Section~\ref{subsec:propagation}), then
develops the completion operator that intercepts that path
(Section~\ref{subsec:completion}), the training objective that keeps clean
forecasting intact (Section~\ref{subsec:dualpath}), the inference procedure
(Section~\ref{subsec:inference}), and the added cost
(Section~\ref{subsec:complexity}). Fig.~\ref{fig:overview} gives the overall
arrangement.

\subsection{Failure Propagation in Channel-Mixed Forecasters}
\label{subsec:propagation}

The accuracy advantage of channel mixing and its exposure to sensor failure come
from the same operator. Consider the pooled core
$\mathbf{s}=\pi(\mathbf{H})$ that an aggregate-and-dispatch layer computes and
redistributes. Under the substitution of Section~\ref{subsec:failure}, a failed
sensor $c\in\mathcal{F}$ still produces an embedding, but that embedding is
$\phi(\mathrm{sub}(\mathbf{x}_c))$ rather than $\phi(\mathbf{x}_c)$. Computing the pooled core over the full set gives
\begin{equation}
\label{eq:corrupt}
\tilde{\mathbf{s}} \;=\; \pi\big(\{\phi(\mathbf{x}_c)\}_{c\in\mathcal{A}}
\cup \{\phi(\mathrm{sub}(\mathbf{x}_c))\}_{c\in\mathcal{F}}\big).
\end{equation}
Every channel receives $\tilde{\mathbf{s}}$, including the healthy sensors in
$\mathcal{E}$ whose own readings were never disturbed. The forecast for a
still-reporting sensor therefore degrades because a different sensor went
dark. Mixing turns each sensor into a dependency of every other one.

Two observations follow, and together they determine the design. The corruption
enters through the embedding, not through the raw signal, as $\pi$ never sees
$\mathbf{x}_c$ and only receives $\mathbf{h}_c$. Also, the quantity the mixing operator
requires from a failed sensor is its embedding $\mathbf{h}_c\in\mathbb{R}^{D}$,
not its $L$ raw readings. Repairing the failure in the space where it acts is
therefore a lower-dimensional problem than reconstructing the signal that would
have produced it, and it is the problem the mixing operator actually poses.

\subsection{Channel-Axis Completion}
\label{subsec:completion}

LCC replaces the embeddings of failed sensors with predictions made from the
surviving ones. Let $\mathbf{m}\in\{0,1\}^{C}$ indicate unavailability, with
$m_c=1$ for $c\in\mathcal{F}$. A learned vector
$\mathbf{q}\in\mathbb{R}^{D}$ marks a channel whose embedding is unknown, and the
masked embedding matrix is
\begin{equation}
\label{eq:mask}
\mathbf{H}^{\mathbf{m}}_{c,:} \;=\;
\begin{cases}
\mathbf{q}^{\!\top}, & m_c = 1,\\[2pt]
\mathbf{h}_c^{\!\top}, & m_c = 0.
\end{cases}
\end{equation}
A completion operator $g$ maps the masked set to predicted embeddings,
$\hat{\mathbf{H}} = g(\mathbf{H}^{\mathbf{m}})$. Because $g$ must infer a
channel's representation from the other channels, it is itself a cross-channel
aggregator; the same aggregate-and-dispatch primitive already present in the
backbone serves this role, which keeps the added machinery to one block. The
completed representation substitutes predictions only where the mask is set,
\begin{equation}
\label{eq:complete}
\mathbf{H}^{\star}_{c,:} \;=\;
m_c\,\hat{\mathbf{H}}_{c,:} + (1-m_c)\,\mathbf{H}_{c,:},
\end{equation}
preserving the original embeddings of surviving sensors exactly.

Training $g$ requires a target for the masked positions. LCC uses the embeddings
themselves, taken as fixed:
\begin{equation}
\label{eq:aux}
\mathcal{L}_{\mathrm{comp}}
= \frac{1}{|\mathcal{M}|}\sum_{c\in\mathcal{M}}
\big\| \hat{\mathbf{H}}_{c,:} - \mathrm{sg}\!\left[\mathbf{H}_{c,:}\right] \big\|_2^2 ,
\end{equation}
where $\mathcal{M}=\{c: m_c=1\}$ and $\mathrm{sg}[\cdot]$ denotes the
stop-gradient. Prediction happens entirely in representation space, with no
decoder back to sensor readings and no reconstruction term, which
distinguishes a joint-embedding predictive objective from masked
reconstruction~\cite{assran2023ijepa,bardes2024vjepa}. The stop-gradient is what
prevents the encoder from making the completion task easy by collapsing
$\phi$ toward representations that are trivially predictable from context.

During training, the mask is sampled rather than observed. Each channel is masked
independently with probability $r$, which controls how much context the completion
operator learns to work from. Small $r$ trains $g$ on near-complete networks and
gives it little practice at heavy loss, while large $r$ forces it to reconstruct
from a sparse surviving set. Section~\ref{sec:setup} gives the value used.

\subsection{Dual-Path Training}
\label{subsec:dualpath}

A single-path arrangement in which the forecasting head always consumes
$\mathbf{H}^{\star}$ costs accuracy on complete inputs. The head then never
observes an incompleted representation during training, making the ordinary
operating condition of the network (all sensors reporting) out-of-distribution at inference. LCC therefore trains both routes through the
same parameters. Writing $\psi$ for the mixing stack and forecasting head, the
objective is
\begin{equation}
\label{eq:loss}
\mathcal{L} \;=\;
\underbrace{\ell\big(\psi(\mathbf{H}),\mathbf{Y}\big)}_{\text{clean path}}
\;+\;
\underbrace{\ell\big(\psi(\mathbf{H}^{\star}),\mathbf{Y}\big)}_{\text{completion path}}
\;+\;
\lambda\,\mathcal{L}_{\mathrm{comp}},
\end{equation}
with $\ell$ the squared error over the forecast horizon. The first term keeps
complete-input forecasting in-distribution, the second trains the route taken
when sensors are dark, and $\lambda$ sets how strongly the auxiliary objective
competes with forecasting for encoder capacity. Both forecasting terms use the
same $\psi$, ensuring that the two routes share every forecasting parameter without requiring a separate failure-mode network.

\subsection{Inference Under Declared Failure}
\label{subsec:inference}

At inference, the mask is no longer sampled. Under assumption~A1, the monitoring
system reports which sensors are dark. Accordingly, $\mathbf{m}$ is set directly from
$\mathcal{F}$, the completion operator fills those positions through
\eqref{eq:complete}, and forecasting proceeds from $\mathbf{H}^{\star}$. When
$\mathcal{F}=\varnothing$, the mask is empty, \eqref{eq:complete} reduces to the
identity, and the forward pass is exactly that of the underlying forecaster,
completion is not invoked and costs nothing. Algorithm~\ref{alg:lcc} states the
procedure.

\begin{algorithm}[!t]
\caption{Latent Channel Completion at Inference}
\label{alg:lcc}
\begin{algorithmic}[1]
\REQUIRE window $\tilde{\mathbf{X}}$, declared failure set $\mathcal{F}$
\ENSURE forecast $\hat{\mathbf{Y}}$
\STATE $\mathbf{h}_c \leftarrow \phi(\tilde{\mathbf{x}}_c)$ for all $c\in\mathcal{C}$
\STATE $m_c \leftarrow \mathbb{1}[c\in\mathcal{F}]$
\IF{$\mathcal{F}\neq\varnothing$}
  \STATE form $\mathbf{H}^{\mathbf{m}}$ by \eqref{eq:mask}
  \STATE $\hat{\mathbf{H}} \leftarrow g(\mathbf{H}^{\mathbf{m}})$
  \STATE $\mathbf{H}^{\star} \leftarrow$ substitution by \eqref{eq:complete}
\ELSE
  \STATE $\mathbf{H}^{\star} \leftarrow \mathbf{H}$
\ENDIF
\STATE $\hat{\mathbf{Y}} \leftarrow \psi(\mathbf{H}^{\star})$
\end{algorithmic}
\end{algorithm}

\subsection{Complexity}
\label{subsec:complexity}

The completion operator adds one aggregate-and-dispatch block over the
$C\times D$ embedding matrix, costing $O(CD^2)$ time and $O(D^2)$ parameters,
independent of the window length $L$ and the horizon $H$. Training adds a second
pass of the mixing stack and head on $\mathbf{H}^{\star}$, bringing the per-step cost
to roughly twice that of the backbone; inference adds only the completion block,
and only when $\mathcal{F}\neq\varnothing$.

The comparison that matters operationally is against repairing the failure in
data space. Reconstructing $|\mathcal{F}|$ sensor windows from
$|\mathcal{A}|$ surviving ones with a linear cross-channel map costs
$O(L\,|\mathcal{A}|\,|\mathcal{F}|)$ per window and stores a $C\times C$
coefficient matrix. LCC's cost does not scale with $L$, while the data-space
route does. Their respective parameter costs are $O(D^2)$ and $O(C^2)$; the smaller of the two therefore depends on
the embedding width relative to the network size.
Section~\ref{sec:results} reports measured parameter counts and latencies for
both.

\section{Experimental Setup}
\label{sec:setup}

\subsection{Monitoring Networks}
\label{subsec:datasets}

Evaluation uses five deployed monitoring networks, summarized in
Table~\ref{tab:datasets}. Three are highway traffic-detector networks from the
California Performance Measurement System, PEMS03, PEMS04 and
PEMS08~\cite{song2020stsgcn}, recording flow at five-minute intervals across
$358$, $307$ and $170$ detectors. The remaining two extend the evaluation beyond
that family: traffic, a larger road-occupancy network of $862$
sensors, and electricity, hourly consumption for $321$
clients~\cite{lai2018lstnet}. All five are public and are used at their released
sampling rates without resampling.

Splits follow the chronological convention established for these
benchmarks~\cite{wu2021autoformer}. The series is divided along time, and
windows are drawn entirely within each split, preventing any window from spanning a split boundary. Because
the split is temporal rather than random, no future observation of a sensor
informs a prediction about its past. Normalization statistics are computed on
the training portion alone.

\subsection{Failure Protocol}
\label{subsec:protocol}

Each network is partitioned once, before any training, into an evaluation set
$\mathcal{E}$ containing half the sensors and a failure pool
$\mathcal{C}\setminus\mathcal{E}$ containing the remainder. The partition is
drawn uniformly at random with a fixed seed and is identical for every method,
every failure rate, and every repetition. Sensors in $\mathcal{E}$ never fail,
and error is computed on $\mathcal{E}$ alone.

Failure episodes are generated by drawing a fraction $p$ of the pool uniformly
at random, with $p\in\{0,0.4,0.8,1.0\}$; $p=1.0$ leaves half the network dark.
The same failure sets are used across methods to ensure paired comparisons.
Failed sensors receive their training-set channel mean over the whole window,
which is the substitution a monitoring pipeline applies when a stream stops.

This design gives the comparison a property, in that a channel-independent
forecaster is unaffected by it. Its prediction for a sensor in $\mathcal{E}$
depends only on that sensor's own history, which never changes; its error is
constant in $p$ by construction. Any variation measured along the failure axis is
therefore attributable to channel mixing rather than to the protocol.

\subsection{Methods Compared}
\label{subsec:baselines}

Four systems are compared, all sharing the same splits, windows, evaluation
channels and failure sets.

SOFTS~\cite{han2024softs} is the channel-dependent backbone, an
aggregate-and-dispatch forecaster whose STAR layer pools the channel embeddings
into a shared core. It receives mean-substituted inputs under failure and is the
model LCC modifies.

PatchTST~\cite{nie2023patchtst} is the channel-independent reference, a
patch transformer that forecasts each sensor from its own history. It anchors the comparison, as the margin between a channel-dependent forecaster and this reference is the accuracy gained from mixing and the quantity whose behaviour under failure this work examines.

Cross-channel ridge imputation repairs failures in data space and is the
strongest comparator for LCC. For each sensor, a ridge predictor is fitted on the
training split to reconstruct that sensor's readings from the remaining sensors;
at inference, the failed windows are reconstructed and passed to an unmodified
SOFTS. Because a whole-window outage leaves no temporal context for the
dead sensor, carry-forward and interpolation are inapplicable, and cross-channel
regression is the operative form of imputation in this regime. The comparison is tightly controlled, as this baseline uses exactly the information available to LCC, but in data space rather than latent space.

LCC is the method of Section~\ref{sec:method}, built on the same SOFTS backbone with identical architecture and optimization settings, allowing the comparison to isolate the completion mechanism rather than
capacity or tuning.

\subsection{Metrics}
\label{subsec:metrics}

Forecast error is mean squared error over the evaluation channels and the full
horizon,
\begin{equation}
\label{eq:mse}
\mathrm{MSE} \;=\; \frac{1}{|\mathcal{E}|\,H}
\sum_{c\in\mathcal{E}}\sum_{t=1}^{H}
\big(\hat{Y}_{t,c}-Y_{t,c}\big)^2 ,
\end{equation}
lower being better. Because absolute error is not comparable across networks with
different scales, results are reported as a relative margin against the
channel-independent reference,
\begin{equation}
\label{eq:advantage}
A \;=\; 100 \cdot
\frac{\mathrm{MSE}_{\textsc{PatchTST}} - \mathrm{MSE}_{\textsc{method}}}
     {\mathrm{MSE}_{\textsc{PatchTST}}} \;\;[\%],
\end{equation}
which we call the mixing advantage. It is positive when a
channel-dependent forecaster beats the channel-independent reference and negative
when mixing has become a liability. Evaluated at $p=0$, it measures how much
cross-channel structure a network offers; evaluated at $p>0$ it measures how much
of that structure survives sensor loss. Both readings use the same quantity, which
is what allows the two to be compared directly.

\subsection{Implementation}
\label{subsec:implementation}

SOFTS uses model width $512$, core width $512$, feed-forward width $512$
and two encoder layers, trained for up to $30$ epochs with early stopping after
three epochs without validation improvement. PatchTST uses patch length
$16$ with stride $8$, model width $128$, $16$ heads, three encoder layers and
feed-forward width $256$, trained for up to $100$ epochs with a patience of ten.
These are the configurations at which both reproduce their published behaviour;
weakening either would make the comparison uninformative. Both use AdamW with
weight decay $10^{-5}$, learning rates $3\times10^{-4}$ and $10^{-4}$
respectively, and gradient-norm clipping at $5$.

LCC adds the completion block and the mask vector to the SOFTS configuration and changes nothing else. The masking probability is $r=0.3$, and
the auxiliary weight is $\lambda=0.1$. The ridge imputer uses regularization
strength $1$, fitted once per network on the training split.

Every configuration is run with two random seeds, which control initialization
and batch ordering; reported values are means over seeds, and intervals where
given are bootstrap percentile intervals over the seed distribution. Experiments
ran on a single NVIDIA GH200 accelerator under PyTorch. Model selection uses
validation error only; the test split is read once per configuration, after
selection. Code, configurations, and seeds are released at \url{\codeurl}.

\begin{table}[!t]
\centering
\caption{Monitoring networks used in the evaluation. Windows are counted after
chronological splitting; the evaluation set $\mathcal{E}$ holds half the sensors
and never fails.}
\label{tab:datasets}
\setlength{\tabcolsep}{4pt}
\begin{tabular}{lrrrrrr}
\toprule
Network & Sensors & Steps & Lookback & Train & Val. & Test \\
\midrule
PEMS03      & 358 & 26\,208 & 96  & 15\,581 & 5\,099 & 5\,099 \\
PEMS04      & 307 & 16\,992 & 96  & 10\,052 & 3\,255 & 3\,256 \\
PEMS08      & 170 & 17\,856 & 96  & 10\,570 & 3\,428 & 3\,429 \\
traffic     & 862 & 17\,544 & 96  & 12\,137 & 1\,709 & 3\,461 \\
electricity & 321 & 26\,304 & 336 & 17\,981 & 2\,537 & 5\,165 \\
\bottomrule
\end{tabular}
\end{table}

\section{Results}
\label{sec:results}

\subsection{The Mixing Advantage Survives Sensor Loss}
\label{subsec:main}

Table~\ref{tab:main} reports the mixing advantage of \eqref{eq:advantage} across
the failure axis on the three traffic-detector networks.

Read the first line of each block as a measurement of how much cross-channel
structure a network offers. With every sensor reporting, channel mixing is worth
$42.9\%$, $62.8\%$ and $39.2\%$ over the channel-independent reference on
PEMS03, PEMS04 and PEMS08. Read down the block to see what sensor loss does to
that advantage. For the unmodified channel-dependent forecaster, it erodes
steadily and, on PEMS03, reverses: at $p=1.0$, where half the network is dark,
SOFTS trails the channel-independent reference by $7.4\%$. Mixing has
stopped paying for itself and become a liability.

The corresponding LCC entries do not move. Across the same range the advantage
goes from $43.1\%$ to $42.0\%$ on PEMS03, from $64.0\%$ to $63.6\%$ on PEMS04,
and from $40.2\%$ to $49.0\%$ on PEMS08. Losing half the network costs LCC at
most $1.0$ percentage point, and on PEMS08 its advantage instead rises by $8.8$;
the unmodified backbone loses between $10.2$ and $50.3$ points on the same data.
Whatever cross-channel structure a network offers on complete inputs, LCC still
has it once half the sensors are gone.

Fig.~\ref{fig:degradation} plots the two trajectories. The gap between them is not a constant offset but widens as sensors fail, reflecting the different ways the two systems respond to failure rather than a fixed difference between them.

On complete inputs, LCC and the backbone are level. The differences at $p=0$ are
$0.2$, $1.2$ and $1.0$ percentage points, which is within the seed-to-seed spread
of either system; the completion path is not invoked when nothing has failed, and
the model reduces to its backbone. Failure invariance therefore comes at no
measurable cost on complete inputs, which matters for deployment because a
monitoring network spends most of its time intact.

\subsection{Completion Beats Data-Space Imputation}
\label{subsec:imputation}

A monitoring system that knows a sensor is dark can also try to reconstruct its
readings and forecast as usual. Table~\ref{tab:imputation} compares that route
against LCC.

Cross-channel ridge imputation is a strong repair. Against the mean substitution
that a pipeline applies by default, it lowers error by $38.5\%$ on PEMS03 and
$20.3\%$ on PEMS04 at full pool failure, and the margin grows with the failure
rate on both networks. Reconstructing dead sensors from live ones evidently
recovers much of what mean substitution destroys.

Fig.~\ref{fig:repair} shows both margins. Completing the latent recovers
more. LCC improves on the ridge pipeline by
$13.9\%$ on PEMS03 and $28.6\%$ on PEMS04 at $p=1.0$, and on both networks the
margin widens monotonically as failures accumulate: $3.2\rightarrow13.9$ and
$3.1\rightarrow28.6$ percent across the range. Both systems draw on the same
information, namely the readings of the surviving sensors, but differ in 
where the repair is applied. Reconstructing a sensor's $L$ readings is a harder
estimation problem than predicting the $D$-dimensional embedding that the mixing
operator actually consumes, and the gap between the two grows precisely as more
sensors need repairing.

\subsection{The Mechanism is Completion, Not Masked Training}
\label{subsec:ablation}

LCC trains with masking and completes at inference. Either could be responsible
for the result, as masking during training is a form of augmentation and might
produce a backbone that simply tolerates corrupted inputs better.
Table~\ref{tab:ablation} separates them by holding the trained model fixed and
changing only what happens at inference. One condition passes the declared
failure set to the completion operator; the other withholds it, causing the same trained model to fall back to mean-substituted embeddings.

Completion accounts for the effect. Withholding it costs $33.4\%$, $37.6\%$ and
$31.1\%$ on PEMS03, PEMS04 and PEMS08 at full pool failure, and the deficit
shrinks with the failure rate, reaching zero at $p=0$. That zero is a
consistency check rather than a result; with no failures the mask is empty, the
substitution in \eqref{eq:complete} is the identity, and the two conditions are
the same computation. Observing exactly $0.0\%$ confirms the two arms differ only
where they are intended to.

Masked training alone therefore does not deliver failure invariance. What it does
is make the completion operator accurate enough to be worth invoking.

\subsection{The Advantage LCC Preserves is the One a Network Has}
\label{subsec:applicability}

Sections~\ref{subsec:main}--\ref{subsec:ablation} concern networks with
substantial cross-channel structure. Table~\ref{tab:applicability} extends the
measurement to two networks with little of it.

On electricity, channel mixing is worth $4.4\%$ over the
channel-independent reference with all sensors reporting. After half the network
goes dark 
LCC holds $4.3\%$; completion
contributes nothing measurable. traffic behaves the same way, with a mixing
advantage of $0.7\%$ on complete inputs and $1.5\%$ retained under failure.

These are the same law evaluated near zero. A forecaster that barely uses
cross-channel information has little exposure when that information is lost, and
correspondingly little for completion to protect. Fig.~\ref{fig:tracking} places all five networks on one axis. Points that
retain their full advantage lie on the diagonal, and the LCC entries sit on it
across the whole range, while the unmodified backbone falls below it by an amount
that grows with how much advantage the network had to lose. The quantity that predicts whether LCC will
help is therefore measurable before the method is deployed by fitting the
channel-dependent and channel-independent forecasters once on complete data and
comparing them. Where the margin is small, an operator can use the simpler
channel-independent model and lose nothing under failure.

\subsection{Behaviour Across Horizons and Partial Outages}
\label{subsec:sensitivity}

Table~\ref{tab:horizon} repeats the main measurement at forecast horizons of
$12$, $24$ and $48$ steps on PEMS03. The mixing advantage itself grows with the
horizon, from $11.7\%$ to $42.9\%$, which is consistent with cross-channel
information mattering more as a sensor's own history is exhausted. Under heavy
failure the backbone retains none of it at two of the three horizons, while LCC
retains $9.9\%$, $26.8\%$ and $42.0\%$ -- between $84\%$ and $111\%$ of what the
network offers intact.

A sensor that fails part-way through a window leaves usable history behind it,
which makes carry-forward substitution available and gives a temporal repair
something to work with. Table~\ref{tab:partial} varies the dark fraction $f$ of
the window. Against last-observation carry-forward, LCC leads by $47.0\%$ when
the outage spans the window and by $9.4\%$ when three quarters of the history
survives; against ridge imputation, the margin stays between $12.3\%$ and
$18.0\%$ throughout. LCC's own error is unchanged across $f$, since it completes
the embedding from the surviving sensors regardless of how much of the dead
sensor's window remains. Brief outages are consequently the regime where the
surviving prefix carries information the current formulation does not use, and
conditioning the completion operator on that prefix is the natural extension.

Masking probability has little influence on the outcome. Varying $r$ over
$\{0.1,0.3,0.5\}$ on PEMS03 changes the advantage under full pool failure by less
than $1.1$ percentage points, suggesting that the method does not require tuning
this setting to the deployment's expected failure rate.

\subsection{Cost}
\label{subsec:cost}

Table~\ref{tab:cost} reports parameter counts and inference latency. LCC adds
$1.2$--$1.3$M parameters to a $3.75$M-parameter backbone, and increases latency
by $8.6\%$ to $69.4\%$ relative to the ridge imputation pipeline depending on
network size. The completion operator is a full aggregation block, whereas the
ridge imputer stores a $C\times C$ coefficient matrix that is comparatively
cheap at these network sizes.

The forecasting error reductions reported in Section~\ref{subsec:imputation} are therefore bought with
computation rather than saved alongside it. For a monitoring network issuing
forecasts on a five-minute cycle, an inference cost of a few milliseconds is not
the binding constraint, and the relevant comparison is against the operational
consequence of forecasts that degrade when detectors fail.

\begin{table}[!t]
\centering
\caption{Mixing advantage over the channel-independent reference,
\eqref{eq:advantage}, as the sensor-failure rate $p$ increases. Higher is
better; negative means channel mixing costs accuracy. $\Delta$\textsc{LCC} is
the change in LCC's advantage from the complete-network condition, computed
before rounding. Means over $n=2$ seeds.}
\label{tab:main}
\setlength{\tabcolsep}{5pt}
\begin{tabular}{llrrr}
\toprule
Network & $p$ & \textsc{SOFTS} & \textsc{LCC} & $\Delta$\textsc{LCC} \\
\midrule
\multirow{3}{*}{PEMS03}
 & 0.0 & 42.9 & \textbf{43.1} & -- \\
 & 0.8 &  8.6 & \textbf{42.9} & $-0.2$ \\
 & 1.0 & $-7.4$ & \textbf{42.0} & $-1.0$ \\
\midrule
\multirow{3}{*}{PEMS04}
 & 0.0 & 62.8 & \textbf{64.0} & -- \\
 & 0.8 & 45.6 & \textbf{63.9} & $-0.1$ \\
 & 1.0 & 35.8 & \textbf{63.6} & $-0.5$ \\
\midrule
\multirow{3}{*}{PEMS08}
 & 0.0 & 39.2 & \textbf{40.2} & -- \\
 & 0.8 & 30.2 & \textbf{42.9} & $+2.7$ \\
 & 1.0 & 29.1 & \textbf{49.0} & $+8.8$ \\
\bottomrule
\end{tabular}
\end{table}

\begin{table}[!t]
\centering
\caption{Error reduction from repairing declared failures, relative to the mean
substitution a monitoring pipeline applies by default. Cross-channel ridge
imputation repairs sensor readings; LCC completes latent representations. Means
over $n=2$ seeds.}
\label{tab:imputation}
\setlength{\tabcolsep}{5pt}
\begin{tabular}{llrr}
\toprule
 & & \multicolumn{2}{c}{Error reduction (\%)} \\
\cmidrule(lr){3-4}
Network & $p$ & Ridge over mean & \textsc{LCC} over ridge \\
\midrule
\multirow{4}{*}{PEMS03}
 & 0.0 &  0.0 &  3.2 \\
 & 0.4 & 11.7 &  4.3 \\
 & 0.8 & 33.3 &  8.0 \\
 & 1.0 & 38.5 & \textbf{13.9} \\
\midrule
\multirow{4}{*}{PEMS04}
 & 0.0 &  0.0 &  3.1 \\
 & 0.4 &  9.8 &  6.7 \\
 & 0.8 & 18.0 & 18.8 \\
 & 1.0 & 20.3 & \textbf{28.6} \\
\bottomrule
\end{tabular}
\end{table}

\begin{table}[!t]
\centering
\caption{Contribution of inference-time completion, isolated by holding the
trained model fixed and withholding the declared failure set, causing the model to fall back to mean-substituted
embeddings. Entries give the error
increase from withholding completion. Means over $n=2$ seeds.}
\label{tab:ablation}
\setlength{\tabcolsep}{6pt}
\begin{tabular}{lrrr}
\toprule
$p$ & PEMS03 & PEMS04 & PEMS08 \\
\midrule
0.0 &  0.0 &  0.0 &  0.0 \\
0.8 & 23.8 & 29.1 & 21.7 \\
1.0 & \textbf{33.4} & \textbf{37.6} & \textbf{31.1} \\
\bottomrule
\end{tabular}
\end{table}

\begin{table}[!t]
\centering
\caption{Mixing advantage on complete inputs and after half the network fails,
for all five monitoring networks. The advantage LCC retains under failure tracks
the advantage the network offers intact. Means over $n=2$ seeds.}
\label{tab:applicability}
\setlength{\tabcolsep}{5pt}
\begin{tabular}{lrrrr}
\toprule
 & & \multicolumn{2}{c}{Advantage (\%)} & \\
\cmidrule(lr){3-4}
Network & Sensors & intact & \textsc{LCC} at $p{=}1$ & Completion (\%) \\
\midrule
PEMS04               & 307 & 62.8 & 63.6 & 37.6 \\
PEMS03               & 358 & 42.9 & 42.0 & 33.4 \\
PEMS08               & 170 & 39.2 & 49.0 & 31.1 \\
electricity          & 321 &  4.4 &  4.3 & $-0.0$ \\
traffic              & 358\rlap{$^{\dagger}$} &  0.7 &  1.5 & 7.1 \\
\bottomrule
\end{tabular}
\begin{minipage}{\columnwidth}
\vspace{2pt}\footnotesize
$^{\dagger}$ traffic is subsampled to 358 sensors to match PEMS03, holding fixed across the comparison.
\end{minipage}
\end{table}

\begin{table}[!t]
\centering
\caption{Mixing advantage across forecast horizons on PEMS03, with all sensors
reporting and after half the network fails. Means over $n=2$ seeds.}
\label{tab:horizon}
\setlength{\tabcolsep}{6pt}
\begin{tabular}{lrrr}
\toprule
 & \multicolumn{3}{c}{Advantage (\%)} \\
\cmidrule(lr){2-4}
Horizon & intact & \textsc{SOFTS} at $p{=}1$ & \textsc{LCC} at $p{=}1$ \\
\midrule
12 & 11.7 & $-8.6$ & \textbf{9.9} \\
24 & 24.2 & 3.3    & \textbf{26.8} \\
48 & 42.9 & $-7.4$ & \textbf{42.0} \\
\bottomrule
\end{tabular}
\end{table}

\begin{table}[!t]
\centering
\caption{Error reduction of LCC over two repairs that use the surviving portion
of a partially dark window, as a function of the dark fraction $f$. PEMS03,
horizon 48, all pool sensors affected. Means over $n=2$ seeds.}
\label{tab:partial}
\setlength{\tabcolsep}{6pt}
\begin{tabular}{lrr}
\toprule
 & \multicolumn{2}{c}{\textsc{LCC} error reduction (\%)} \\
\cmidrule(lr){2-3}
Dark fraction $f$ & over carry-forward & over ridge \\
\midrule
1.00 & \textbf{47.0} & 13.9 \\
0.75 & 27.6 & 15.0 \\
0.50 & 20.7 & 12.3 \\
0.25 &  9.4 & \textbf{18.0} \\
\bottomrule
\end{tabular}
\end{table}

\begin{table}[!t]
\centering
\caption{Deployment cost of repairing declared failures. Latency is the median
over test batches at full pool failure; the ridge pipeline additionally stores a
$C\times C$ coefficient matrix.}
\label{tab:cost}
\setlength{\tabcolsep}{5pt}
\begin{tabular}{lrrr}
\toprule
Network & Ridge params (M) & \textsc{LCC} params (M) & $\Delta$ latency (\%) \\
\midrule
PEMS03 & 3.88 & 5.07 & $+69.4$ \\
PEMS04 & 3.85 & 5.07 & $+8.6$  \\
PEMS08 & 3.78 & 5.07 & $+9.8$  \\
\bottomrule
\end{tabular}
\end{table}

\section{Discussion}
\label{sec:discussion}

\subsection{Deployment Implications}
\label{subsec:deployment}

For an operator running a detector network, the results change what happens
during an outage rather than what happens on a good day. A coupled forecaster
selected on complete-network benchmarks carries an exposure that those benchmarks
do not capture. On PEMS03, it forecasts worse than an uncoupled model once half the
detectors are dark, making the architecture chosen for its accuracy the wrong choice when the network is degraded. 
Completion removes the reversal without requiring a second,
separately trained model to maintain, since the completion operator
is trained jointly with the forecaster and shares its parameters.

The criterion for whether to deploy it is cheap to evaluate. Across the five
networks studied, the advantage LCC retains under failure tracks the advantage
the network offers intact, from $0.7\%$ to $62.8\%$. An operator can therefore
fit a coupled and an uncoupled forecaster once on complete data and read the
margin between them. Where the margin is large, coupling is worth protecting and
completion protects it. Where the margin is small, the uncoupled model is the simpler
choice and loses nothing under failure. This decision needs no failure data and
no outage history.

There is a further consequence for the operator supervising the pipeline. 
A forecaster whose accuracy ranking against a simpler alternative reverses during an outage is one whose reliance cannot be calibrated from its ordinary behaviour. Calibrated reliance is what trust management frameworks for human-AI teams require of a deployed component, with performance that is continuously measurable and behaviour that remains predictable as conditions change~\cite{hou2025impacts,hou2023challenges}.
Failure invariance supplies the
second property directly, and the criterion above supplies the first, since the
quantity that predicts behaviour under outage is measured before any outage
occurs. This paper does not model operator trust, and evaluating whether these
guarantees change how operators actually rely on such a system would require a
human-subjects study we have not conducted.

Assumption~~\textbf{A1} sets the boundary of applicability. LCC requires knowing which
sensors are dark, which monitoring systems track independently of the forecaster
and which is available whenever a stream stops transmitting. It does not apply to
sensors that fail silently while continuing to report plausible values; detecting
that condition is a separate problem, and a forecaster given a wrong failure mask
would complete healthy channels it should have left alone.

\subsection{Threats to Validity}
\label{subsec:threats}

\emph{Construct Validity.} Forecast error on the evaluation channels measures
what an operator loses on the sensors still reporting, which is the quantity the
paper is about. It does not capture downstream decision quality, as whether a
$40\%$ error reduction changes a ramp-metering action depends on the control
policy, and that link is not evaluated here. The mixing advantage of
\eqref{eq:advantage} is a ratio against a channel-independent reference rather
than an absolute error, chosen to enable comparisons across networks with different scales, and it inherits whatever weaknesses that reference
has on a given network.
\emph{Internal Validity.} The evaluation channels are fixed before training and
never fail, ensuring that error is computed on the same sensors at every failure rate and for every method. The channel-independent reference is unaffected by the failure
protocol by construction, which provides a control, as measured variation along the
failure axis cannot come from the protocol itself. Failure sets are drawn once
per rate and shared across methods, making the comparisons paired. Model
selection reads validation error only.

\emph{External Validity.} The evidence covers five public monitoring networks,
three of them from one traffic-detector family, at horizons from $12$ to $48$
steps. The failure model removes sensors uniformly at random from a designated
pool; real outages can be spatially clustered, and although
Section~\ref{subsec:sensitivity} varies how much of a window is dark, it does not
vary which sensors fail together. Networks whose sensors are coupled through
mechanisms other than spatial proximity may behave differently.

\emph{Conclusion Validity.} Reported values are means over two seeds per
configuration, and the intervals quoted are bootstrap percentile intervals over
that distribution. The effects the paper rests on are large relative to the
observed seed spread, as the difference between a coupled model retaining $42.0\%$
and losing to an uncoupled reference at $-7.4\%$ is not a marginal one. Smaller
differences, in particular the comparison on complete inputs where LCC and its
backbone are level, should be read as showing no measurable difference rather
than as establishing an ordering.

\subsection{Limitations}
\label{subsec:limitations}

Brief outages leave information unused. LCC completes a channel's embedding from
the surviving sensors regardless of how much of that channel's own window
survived; its error therefore remains unchanged across dark fractions. When most of a
window remains, carry-forward substitution recovers ground on it, from a $47.0\%$
deficit at a full outage to $9.4\%$ when three quarters of the history survives.
Conditioning the completion operator on the observed prefix, rather than
replacing the embedding outright, would let the method use what remains; this is
a change to the completion operator's input rather than to the objective.

The completion operator adds computation. It contributes $1.2$--$1.3$M
parameters and increases inference latency relative to reconstructing readings
with a linear cross-channel map, by $8.6\%$ to $69.4\%$ depending on network
size. On a five-minute forecast cycle this is unlikely to bind, but on
higher-rate monitoring or embedded deployment it might, and a lighter completion
operator would be worth
examining.

Sensor coupling is treated as a single undifferentiated quantity. The method
protects whatever advantage coupling provides without identifying which sensors
supply it, offering no guidance on where to prioritize repair when
maintenance capacity is limited. Attributing the retained advantage to individual
surviving sensors would connect the method to inspection scheduling, which is
where a reliability engineer would want it.

Finally, the evidence concerns simultaneous failure of sensors drawn at random.
Outages in a real network may be correlated with the traffic conditions that make
forecasting difficult, in which case failures and forecast difficulty are not
independent as they are here. Evaluating on recorded outage logs, rather than
synthetic failure sets, would test that.

\section{Conclusion}
\label{sec:conclusion}

We presented Latent Channel Completion, a channel-axis joint-embedding
predictive objective that reconstructs missing sensors' latent representations
rather than their values. The design followed from locating where sensor failure
damages a coupled forecaster, with a substituted reading for a dark sensor becoming a corrupted embedding, the shared representation computed across all sensors
inheriting that corruption, and forecasts degrading for sensors that never failed.
Because the coupling operator consumes embeddings rather than readings, supplying
the embedding directly addresses the failure where it acts. On three highway detector networks, the accuracy advantage of coupling became
insensitive to sensor loss. After half of each network went dark, that advantage
changed by $-1.0$, $-0.5$ and $+8.8$ percentage points under LCC, against losses
of $10.2$ to $50.3$ points for the unmodified forecaster, which on one network
fell behind an uncoupled reference entirely. Completing the latent also outperformed reconstructing the readings, with error falling by a further $13.9\%$ and $28.6\%$ against cross-channel ridge imputation drawing on the same surviving sensors. An ablation holding the trained model fixed showed the effect came from
completion at inference rather than from masked training, and on complete inputs
LCC matched its backbone. The benefit tracked how much a network gains from coupling in the first place,
measured across five networks spanning $0.7\%$ to $62.8\%$, which gives an
operator a criterion evaluable on complete data before any outage occurs. What
remains open is the brief-outage regime, where a partially dark window retains
history that the current completion operator discards.

\bibliographystyle{IEEEtran}
\bibliography{refs}

\end{document}